\documentclass[conference]{IEEEtran} 
\IEEEoverridecommandlockouts
\usepackage{cite}
\usepackage{amsmath,amssymb,amsfonts,amsthm}
\usepackage{mathtools}
\usepackage{algpseudocode}
\usepackage{graphicx}
\usepackage{textcomp}
\usepackage{color}
\usepackage[ruled,vlined]{algorithm2e}
\usepackage{multirow}
\usepackage{import}
\usepackage{bm}
\usepackage{caption}
\usepackage{subcaption}
\usepackage{comment}
\usepackage{soul}

\def\BibTeX{{\rm B\kern-.05em{\sc i\kern-.025em b}\kern-.08em
    T\kern-.1667em\lower.7ex\hbox{E}\kern-.125emX}}
\makeatletter
\def\endthebibliography{%
	\def\@noitemerr{\@latex@warning{Empty `thebibliography' environment}}%
	\endlist
}
\makeatother

\DeclareMathOperator*{\argmin}{arg\,min}

\ifCLASSINFOpdf
\else
   \usepackage[dvips]{graphicx}
\fi
\usepackage{url}

\usepackage{graphicx}

\begin{document}

\title{Diff-GO: Diffusion Goal-Oriented Communications to Achieve Ultra-High Spectrum
Efficiency }


\author{\IEEEauthorblockN{Achintha Wijesinghe$^1$, Songyang Zhang$^2$,~\textit{Member, IEEE}, Suchinthaka Wanninayaka$^1$,\\ Weiwei Wang$^1$, and
Zhi Ding$^1$, \textit{Fellow,~IEEE}}
\IEEEauthorblockA{$^1$University of California, Davis, CA, USA \\
$^2$University of Louisiana at Lafayette, LA, USA}}


\markboth{ }
{Shell \MakeLowercase{\textit{et al.}}: Bare Demo of IEEEtran.cls for IEEE Journals}
\maketitle

\begin{abstract}
The latest advances in artificial intelligence (AI) present 
many unprecedented opportunities to achieve much improved
bandwidth saving in communications. Unlike conventional communication systems focusing on packet transport, rich datasets and
AI makes it possible to efficiently transfer 
only the information most critical to
the goals of message recipients. One of
the most exciting advances in generative AI known as
diffusion model presents a unique opportunity for designing ultra-fast 
communication systems well beyond
language-based messages. This work presents an
ultra-efficient communication design 
by utilizing generative AI-based
on diffusion models as a specific example of
the general
goal-oriented communication framework.  To better control 
the regenerated message at the receiver output, our diffusion system design
includes a local regeneration module 
with finite dimensional noise latent.
The critical significance of noise latent control and sharing residing on our Diff-GO is the ability to introduce the
concept of ``local generative feedback" (Local-GF), which
enables 
the transmitter to monitor the quality and
    gauge the quality or accuracy of the message
    recovery at the semantic system
    receiver. To this end, we propose a new low-dimensional noise space for the training of diffusion models, which significantly reduces the communication overhead and achieves satisfactory
message recovery performance. Our experimental results demonstrate that the proposed  noise space and the diffusion-based generative model achieve ultra-high spectrum efficiency
and accurate recovery of transmitted image signals. 
By trading off computation for bandwidth efficiency (C4BE), 
this new framework provides an important avenue 
to achieve exceptional computation-bandwidth tradeoff.
\end{abstract}


\IEEEpeerreviewmaketitle

\section{Introduction}
In the ever-evolving landscape of communications and networking, novel technologies, such as the sixth generation (6G)~\cite{6G_1} and autonomous driving, are expected to emphasize the need for effective and efficient methods of spectrum utilization by incorporating the rapidly growing artificial intelligence (AI).  In recent years, the field of generative AI has witnessed a paradigm shift with the advent of diffusion models~\cite{DDPM}. These models, built upon the foundations of deep learning and neural networks, have revolutionized the way we process and understand language and visuals ushering in a new era for bandwidth-efficient communication. These advances present opportunities for the study of semantic communication~\cite{6G_2} by focusing on the delivery of a message's ``meaning" rather than the original signal manifestation.
Semantics-based communication has proven to be a promising direction toward goal-oriented communication by having a plethora of far-reaching real-world applications with the prevailing computational-sound smart transmitters. For example, a system to communicate text information
semantically with a customized loss function was proposed in~\cite{text}. Other applications also include UAV communication, remote image sensing and fusion, intelligent transportation, and healthcare~\cite{summary}.

The \textit{goal} of transferring semantic meaning in communication is only a special case of a general framework of \textit{goal-oriented} communications. Semantic communications assume that the goal is to deliver the meaning rather than the exact fully reconstructed message~\cite{sem_1} to the recipients, thereby enabling efficient information transfer. 
However, in a more broad context, the designers are more 
interested in what the end goal of information transfer is, beyond the mere semantic meaning. For example, if the goal of data collection is to facilitate the model training of autonomous driving, the autonomous vehicle (AV) understands its surroundings by processing images from networked cameras, where information
such as the depth, direction, or size of surrounding objects are more critical to AV decision making than the colors or the brands of adjacent cars. Therefore, in this goal-oriented communication example,
full resolution image is redundant, whereas a simple semantic description of the camera scene is insufficient. 
Generally, goal-oriented communications rely on 
local computation
engine to
determine what critical information should be transmitted and at what accuracy. Bandwidth efficiency can be gained from computation on the transmitter side to outperform classical communication approaches. In many applications including autonomous driving, power transceivers with sufficient
computation power is affordable to achieve the 
desired computation-bandwidth tradeoff without compromising
the specific downstream
goals for communications, such as decision making.
The improved efficiency, link reliability, user quality of experience, and smoother cross-protocol communications justify the strong role of  computation for bandwidth efficiency (C4BE)
designs in future communications. 

Some recent works have started to consider AI-enabled frameworks, integrating many concepts in AI, such as machine learning, causal reasoning, and minimum description length theory~\cite{end-to-end}. One interesting diffusion-based encoder proposed in~\cite{babarosa} aims to combat channel noise 
by sharing a pre-trained diffusion model is used for information generation. Modern wireless systems, however, rely on techniques such as hybrid ARQ to combat packet errors or losses and do not involve noise effects directly onto the raw messages. 
Another work uses masked vector quantized-variational autoencoder (VQ-VAE) for a goal-oriented communication setup~\cite{10101778}. Other typical AI-enabled communication schemes also include auto-encoders~\cite{9814566} joint task and data-oriented semantic communications~\cite{huang2023joint}, semantic multi-modal data
systems \cite{zhang2022unified}, and semantic channel coding \cite{9852388}. Due to the page limits, interested readers could refer to \cite{6G_1, sem_1} for a more complete survey on semantic communications as a special case of Goal-Oriented (GO) communication systems.


\begin{figure*}[htbp]
\centering
\begin{minipage}[t]{0.75\linewidth}
\centering
\includegraphics[width=1\columnwidth]{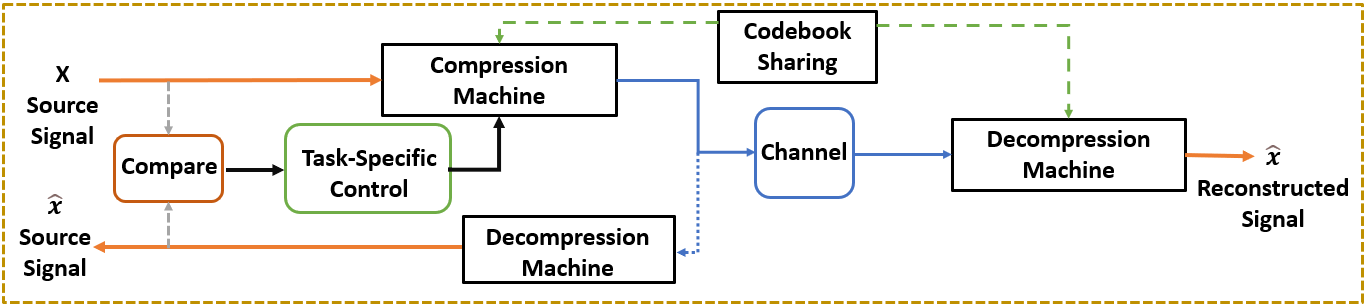}
\caption{Proposed semantic communication framework: The overall diagram;}
\label{fig_1_1}
\end{minipage}
\hspace{3cm}
\vspace{-0.5cm}
\begin{minipage}[t]{0.75\linewidth}
\centering
\includegraphics[width=1\columnwidth]{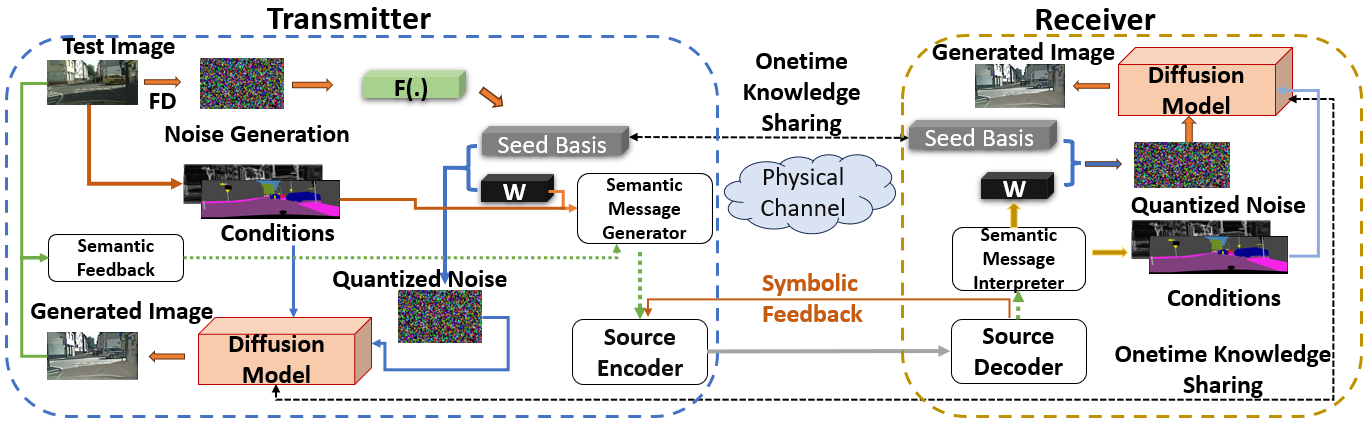}
\caption{Proposed semantic communication framework: The detailed mechanism. } \label{fig_1_2}
\end{minipage}
\end{figure*}


Despite the successes, the existing approach to semantics-based frameworks to enhance bandwidth utilization faces some obstacles. Firstly, the structure of conventional AE-based architecture limits the flexibility and ability of semantic regeneration, where its integration with the emerging generative learning models might be more informative and effective. Secondly, the data link is often designed in one direction and no consideration is given for quality of service (QoS) control, especially for diffusion-based architecture \cite{babarosa}. Due to the random nature of diffusion models, sharing constraint conditions alone has no control over the generated output accuracy. As a result, the transmitter is unaware of the reconstructed output on the receiver side. On the other hand, using a random noise latent to generate the output may not guarantee the closest representation of the input information, especially in images. Moreover, the questions
of how to reduce the communication overhead and how to fully leverage the flexibility of generative learning models remain open.

To tackle the aforementioned practical issues in existing designs, we propose a novel \textit{Diff}usion-based, \textit{G}oal-\textit{O}riented  (Diff-GO) communication framework utilizing generative AI with local generative feedback (local-GF), where a new lightweight low dimensional noise space is proposed for
the training of diffusion models.
Our proposed Diff-GO communication framework aims to reduce the
communication overhead and implement effective information regeneration to satisfy the
required goal-oriented QoS (GO-QOS)at the receiver by designing  a novel
local generative feedback (local GF) at the transmitter. Our contribution can be summarized as follows:

\begin{itemize}
    \item We propose a novel C4BE design principle and
    a Diff-GO communication architecture. Our system design
    employs local GF for GO-QOS
    control.
    \item We propose a new training approach for diffusion models. Specifically, we introduce a noise space mapping which enables noise latent sharing between the transmitter and the receiver at a very low communication cost.
    \item Through rigorous tests, we demonstrate the efficacy of Diff-GO as an effective C4BE design. 
\end{itemize}

\section{Overall Architecture} 
\label{sec:system_model}

In this section, we first present our novel 
communication architecture for task-specific semantics based on the diffusion model as illustrated in Fig.~\ref{fig_1_1}. 

\subsection{Exemplary Application in Autonomous Driving}
In this work, as an example, we consider autonomous driving as an exemplary application and focus on the transmission of city street images for ``smart driving". 
To ensure the collaboration and federation among different autonomous cars in smart driving, efficient communication on street traffic conditions collected from cameras or sensors plays an important role in identifying objects and making safe decisions. In particular, it is paramount to capture the correct road signs and patterns in autonomous driving, where the edge map and segmentation map contain rich go-oriented (GO) information for driving intelligence. For convenience, we will introduce our overall Diff-GO communication architecture in the background of autonomous driving. Note that, beyond autonomous driving, our proposed framework can be easily deployed in any Diff-GO communication scenario with suitable use-case conditions.

\vspace{-0.18cm}
\subsection{Deployment and Local Generation At Transmitter Side}
To leverage generative learning in Diff-GO communication, we investigate the utilization of diffusion models.
For communication efficiency, we propose a novel quantized noise space rather than the continuous noise space spanned from the classical forward diffusion, inspired by the idea of introducing a local generative to eliminate any ambiguity introduced by the nature of the Diff-GO.
In many proposed semantic communication frameworks such as~\cite{babarosa}, the information generated on the receiver side is unknown to the transmitter. To alleviate this issue, we propose a novel low-dimensional (low-DIM) noise space spanned by a linear combination of known $n$ number of seeds and share the corresponding weights ($w$) with the receiver.

As depicted in Fig.~\ref{fig_1_2}, the transmitter is equipped with a pre-trained diffusion model. The diffusion model is pre-trained on a pre-selected noise basis spanned from a known set of $n$ number of seeds. We will elaborate on the low-DIM noise space in Section~\ref{ssec:noise_q}. Our transmitter pipeline starts with noise generation. We first utilize classical forward diffusion to derive the noise latent of any given image. With the derived noise latent, we project the noise latent to the low-DIM noise space and find the optimal weight vector $\mathbf{W}$, which will be presented in Section~\ref{ssec:noise_q} in detail. Simultaneously, we extract conditions from the given image. For a task such as autonomous driving, we propose to use a segment map and edge map to enhance the generation quality of the diffusion model. 

To further reduce the communication overhead, we propose a hierarchical approach to send the most significant $n_i$ weights $n_1 < n_2 < , \dots , n_p< n$ to the receiver. These $n_1, \dots, n$ are predefined. Subsequently, we iterate through each $n_i$ and generate a low-DIM noise representation of the original noise latent by linearly combining the selected $n_i$ weights with the corresponding noise basis generated from seeds. Next, the low-DIM noise latent is fed through the diffusion model for denoising by conditioning on the extracted condition. The output from the diffusion model is sent for evaluation. The evaluation is done by our proposed local generative feedback block. We further illustrate this block in Section~\ref{ssec:feedback}. If the GO-QOS score is adequate for the downstream task, we pick the corresponding weights for the diff-GO message generation followed by transmitter-level encoding and communicate with the receiver through the physical channel.
\vspace{-0.25cm}
\subsection{Data Transmission and Regeneration at Receiver Side}

As in any classical communication system, we can apply feedback and error correction schemes to evaluate the validity of the received information. We adopt a semantic message interpreter on the receiver side to uncover the sent $\mathbf{W}$ and diffusion conditions. We combine the seed basis and $\mathbf{W}$ linearly to recover the best noise latent and feed it to the diffusion model along with the diffusion conditions. This process guarantees to generate the exact image generated on the transmitter side.

\vspace{-0.2cm}
\section{Methodology} \label{sec:method}
\subsection{Diffusion Models}
\label{subsect:diffusion}

Diff-GO pivots around goal-oriented diffusion models. In this part, we first briefly introduce the structure of diffusion models \cite{DDPM}.
In general,
the training of a diffusion model consists of two processes: 1) forward diffusion and 2) backward diffusion. 
Let any data point be denoted by $x_{0} \sim q(x_0) $ and $\{x_1, \cdots,x_T\}$ represent latent with similar dimension to $x_0$. 

In the forward diffusion, a given data point is used to learn a noisy latent by iterative noise addition. In this process, the
posterior $q(x_{1:T} |x_0)$ is determined as a Markov chain that
gradually adds Gaussian noise to the given data point. For example, in autonomous driving, noises will be added to the street image from cameras at each step.
A variance schedule $\{\beta_1, . . . , \beta_T\}$ is used at each iteration~\cite{DDPM} as follows.
\vspace{-0.2cm}
\begin{align}
   q(x_{1:T}|x_0) = \prod_{t=1}^{T} q(x_t|x_{t-1}),\\
   q(x_t|x_{t-1}) = \mathcal{N}(x_t;\sqrt{1-\beta_t}x_{t-1},\beta_t\mathcal{I}).
    \label{eqn:fwd}
\end{align}

In the backward diffusion process, a neural network is utilized as a denoising auto-encoder to learn the added noise in each forward process to characterize the joint distribution $p_{\theta}(x_{0:T})$. The reverse process is modeled as a Markov chain and the aforementioned auto-encoder is expected to learn the Gaussian transitions. This transitions starts at $p(x_T) = \mathcal{N}(x_T;\mathbf{0},\mathbf{\mathcal{I}})$ and follows the tranisition steps below.
\vspace{-0.2cm}
\begin{align}
   p_{\theta}(x_{0:T}) = p(x_T) \prod_{t=1}^{T} p_{\theta}(x_{t-1}|x_t),\\
   p_{\theta}(x_{t-1}|x_{t}) = \mathcal{N}(x_{t-1};\nu_{\theta}(x_t,t),\Sigma_{\theta}(x_t,t)).
    \label{eqn:bwd}
\end{align}

\subsection{Go-Oriented Diffusion Model}
\label{ssec:semantic_diffusion}
The promise of low communication cost go-oriented (GO) communication is viable due to the amazing properties of diffusion models. In this part, we introduce the design of Diff-GO communication systems.

Since diffusion models provide the freedom of conditioning using GO-QOS, we could modify the reverse process (backward diffusion) with a given condition $y$ in conditional diffusion models, recalculated by,
\vspace{-0.4cm}
\begin{align}
   p_{\theta}(x_{0:T}|y) = p(x_T) \prod_{t=1}^{T} p_{\theta}(x_{t-1}|x_t,y),\\
   p_{\theta}(x_{t-1}|x_t,y) = \mathcal{N}(x_{t-1};\nu_{\theta}(x_t,y,t),\Sigma_{\theta}(x_t,y,t)).
    \label{eqn:bwdc}
\end{align}

As introduced in the work~\cite{babarosa}, it is possible to share communication-friendly conditions with the receiver side and generate a possible GO representation of the input data, such as the city images in autonomous driving, by starting with any random noise. However, such an approach has no guarantee that the reconstructed image is accurate in terms of GO-QOS since some sampled noise latent vectors of the learned data latent may not map back to the original space and fail to represent the original image distribution, resulting in generating an image without any valuable information. Fortunately, a conditional diffusion model provides 2 degrees-of-freedom to alter the regeneration to ensure GO-QOS consistency: 1) alter the generated image by changing the input noise latent; or 2) use the conditions to manipulate the model output. 

\begin{figure}[t]
\centering
\begin{minipage}[t]{0.48\linewidth}
\centering
\includegraphics[width=1\columnwidth]{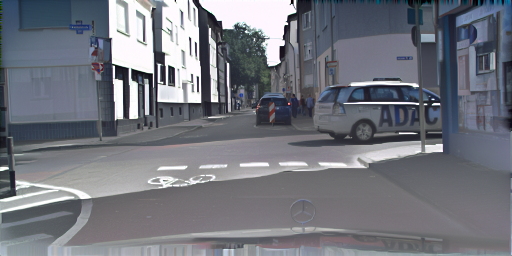}
\caption{Our receiver model output. }
\label{fig:ours}
\end{minipage}
\hfill
\begin{minipage}[t]{0.48\linewidth}
\centering
\includegraphics[width=1\columnwidth]{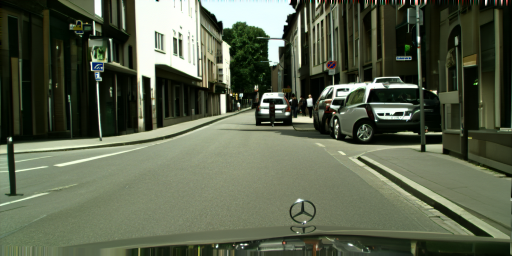}
\caption{Output from the system of~\cite{babarosa}} \label{fig:babarosa}
\end{minipage}
\vspace{-0.3cm}
\end{figure}

To leverage the critical information, we use the edge map as conditions, generated using Canny edge detection from the segmentation map. Compared to the generated images in the ~\cite{babarosa} which only utilize the segmentation map of street images, we can preserve essential information such as road signs by introducing edge maps as conditions, as depicted in Fig~\ref{fig:ours} and Fig.~\ref{fig:babarosa}. 
Despite the successes of the semantic diffusion model, only
sharing the conditions with the receiver still cannot guarantee the quality of regenerated data. Moreover,
the transmitter side cannot be aware of the reconstructed output on the receiver side without any feedback. To alleviate these downsides, we now introduce local generative feedback with a proper choice of noise latent vectors next.

\begin{figure}[t]
\centering
\includegraphics[width=0.9\columnwidth]{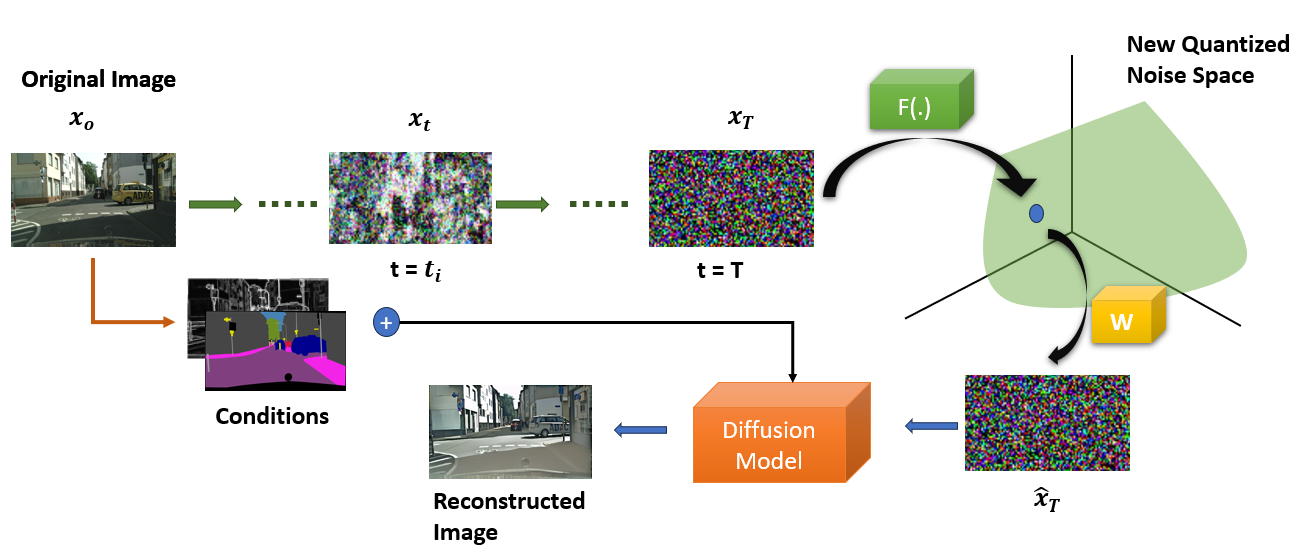}
\caption{Proposed Forward Diffusion: We perform classical forward diffusion up to $t=T$ and map $x_T$ to a new space spanned by $N$ number of known noise vectors. These vectors are defined by a seed value (a numerical value). The diffusion model is then trained on the new projected noise space }
\vspace{-0.4cm}
\label{fig:fig2}
\end{figure}

\begin{algorithm}[h]
    \caption{Diff-GO Communication Training}
    \label{alg1}
    \DontPrintSemicolon
    \LinesNumbered
    \SetNlSty{}{}{:}
    \KwIn{Training data: $\mathcal{D}$}
    \KwIn{Number of vectors in noise basis: $N$}
    \KwOut{Trained diffusion model: $D$}
    \KwOut{Seed set: $\mathcal{S}$}
    Initialize $\mathcal{S}$ randomly: $\mathcal{S} \xleftarrow{} \{s_1, \ldots, s_N \}$\\
    \While{ $D$ is not converged }{
        \For{each $d \in \mathcal{D}$}{
            $x_t \xleftarrow{} $ output of FWD(d)\\
            $m_t \xleftarrow{} $ GO-QOS for $d$\\
            \If {$t == T$} {Solve Eq.~(\ref{eqn:f}) $w \xleftarrow{} F(x_T, \mathcal{S})$ \\
            $\hat{x_T} \xleftarrow{}  \sum_{i=1}^{N}w_i \times s_i $\\
            $x_t \xleftarrow{} \hat{x_T}$}
        Train $D$: $D \xleftarrow{}$ BWD($x_t,m_t$)
        }  
    }
Share $D$ and $\mathcal{S}$ with the receiver
\end{algorithm}\vspace{-0.3cm}

\subsection{Noise Quantization}
\label{ssec:noise_q}
To ensure accurate feedback, we first need to identify the best possible noise latent for a given image. Here, We argue that the best possible noise latent for any given image shall be characterized by the forward diffusion process. This is because the diffusion process is an encoder-decoder structure, where the forward diffusion encodes a given image to a noise latent of the same dimensions of the image, and the backward diffusion decodes or denoises it. Therefore, intuitively, a well-trained diffusion model is capable of denoising a noise latent derived from the forward diffusion back to the original image distribution. Therefore, at the inference, we propose to use the forward diffusion to map a given image to its noise latent. However, if we have to share the derived noise directly with the receiver side, it serves no good since the noise is not compressible. To address this, we propose a new mapping from the noise space of the forward diffusion to a predefined new noise space. Let $s_1, s_2, \cdots, s_n$ be randomly selected $n$ number of noise seeds and $N_1, N_2, \cdots , N_n$ be the corresponding noise latents generated by each noise seed from a Gaussian distribution. We define a new quantized linear noise space $\mathcal{N}_q$ by the span denoted as
${Span}(N_1, N_2, \cdots, N_n)$.

As illustrated in Fig~\ref{fig:fig2}, we first follow a typical forward and backward diffusion process except for the case when the step $t = T$. Every time, the step hits $t = T$ we learn a function $F(.)$ which maps the noise latent at $t=T$ ( $N_T$ ) to noise space $\mathcal{N}_q$. i.e.,
   $F: N_T \to \hat{N_T} \in \mathcal{N}_q$.
    
Suppose that $w_1, \cdots, w_n$ represents a set of weights. The function $F(.)$ can be obtained as follows.
\vspace{-0.33cm}
\begin{align}
   F = \argmin_{w_1, \cdots, w_n} || N_T - \sum_{i=1}^{n}w_i \times N_i ||^2
    \label{eqn:f}
\end{align}
We solve the above optimization problem for each $N_T$ starting with the same weights $w_1, \cdots, w_n$ and use gradient descent to solve the optimization problem in Eq.~(\ref{eqn:f}). This approach allows us to train our diffusion model in a known quantized noise space. The main advantage of this approach is that we can use the forward process to derive the best possible noise and represent it by only $n$ number of weights which is a highly compressed representation of the noise latent. Then, the weights can be sent to the receiver side for reconstruction of the exact noise for backward diffusion. 

\subsection{Local Generative Feedback}
\label{ssec:feedback}
With the quantized space, we can represent the noise with respect to a known basis, where 
we have the freedom of representing any derived noise latent in a highly compressed representation using $n$ floating points. 
The new representation therefore allows us to generate the exact information that the receiver is going to generate even before sharing the information with the receiver. This allows us to validate the reconstruction with the original image using goal oriented QoS (GO-QOS) metrics, such as Fréchet inception distance (FID) score~\cite{FID}, Learned Perceptual Image Patch Similarity (LPIPS)~\cite{LPIPS}, Segment Anything Model  (SAM) score~\cite{SAMscore}, or any downstream task, which will be further discussed in Section \ref{sec:simulations}. This approach significantly differs from existing semantic feedback since we do not wait for the receiver reports (e.g. car crash). Therefore, it provides communication efficiency.
Agorithm~\ref{alg1} and Algorithm~\ref{alg2} summarize the training and the inference of the proposed method.
\vspace{-0.1cm}

\section{Numerical Experiments} \label{sec:simulations}


We now present the performance of our Diff-GO communication system against existing works. We evaluate all the models on the Cityscape dataset~\cite{cityscape} for autonomous driving. 

\vspace{-0.35cm}
\begin{algorithm}[h]
    \caption{Message Inference}
    \label{alg2}
    \DontPrintSemicolon
    \LinesNumbered
    \SetNlSty{}{}{:}
    \KwIn{Inference image: $\mathcal{I}$}
    \KwIn{GO-QOS Threshold: $\tau$}
    \KwIn{GO-QOS metric: $\mathcal{M}$}
    \KwOut{Weights set $w$}
    Initiate a set $\mathcal{P}$ with the different number of important basis vectors to consider $\mathcal{P} \xleftarrow{} {n_1,n_2, \dots, n_q, n }$
       
    $x_t \xleftarrow{} $ output of FWD($\mathcal{I}$)\\
    $m_t \xleftarrow{}$ meaning of $\mathcal{I}$ \\

    Solve Eq.~(\ref{eqn:f}) $w \xleftarrow{} F(x_T, \mathcal{S})$ \\

    \For{each $n_i \in \mathcal{P}$}{
        $wT \xleftarrow{} $ sort from max to min ($|w|$)[$n_i$]\\
        $\hat{w} \xleftarrow{} $ [0 for $w_i$ in w if $|w_i|  < wT$ else $w_i$ ] \\
        $\hat{x_T} \xleftarrow{}  \sum_{i=1}^{n_i}\hat{w}_i \times s_i $\\
            $x_t \xleftarrow{} \hat{x_T}$\\
            $I_i \xleftarrow{}$ BWD($x_t,m_t$)\\
            GO-QOS score $\xleftarrow{} \mathcal{M}(\mathcal{I},I_i)$\\
            \If {$GO-QOS \ score \leq \tau$} {Share $\hat{w}$ and $m_t$ }

        } 
        Share $\hat{w}$ and $m_t$ 
\end{algorithm}\vspace*{-3mm}

\vspace{-0.33cm}
\subsection{Quality of Reconstructed Receiver Message}
First, we evaluate the quality of the reconstructed images by different methods, such as generative Semantic Communication (GESCO)~\cite{babarosa} and original diffusion (OD). In OD, we use guided diffusion~\cite{guided_diffusion} and use the semantic map and edge map as the conditions, where we share the noise latent extracted from the forward diffusion process. In diffusion with random noise (RN), we use a random noise latent to generate results without sharing the noise latent derived from the forward diffusion in Diff-GO. The evaluation is done after training all the models in 250000 steps.
For the evaluations, we rely on GO-QOS measurement and avoid any pixel-wise measures, such as mean square error (MSE), in view of communication goals. For example, the same car having different colors reports a higher MSE value even though they are the same 
for decision making. FID score enables us to evaluate the generated images with respect to human inception at the feature level. Lower FID values represent better reconstruction quality. 
The results are presented Table~\ref{tab:table1} 
From the results, we can see that the model OD has the lowest FID. The reason behind this observation is that we are sharing the entire noise latent with the receiver side. However, as we mentioned, sharing the noise latent is not feasible and the communication cost is even higher compared with image sharing due to the incompressibility of the noise latent. We also observe the effect of the edge map as a condition when we compare it with the higher FID in the GESCO and OD methods. Similar observations can be made for LPIPS as well. For the SAMSR and SAMSS, presented in Table~\ref{tab:table3}, DIff-GO 100 and OD perform closely. We see an increment in performances as $n$ increases in Diff-GO.
In the last column of Table~\ref{tab:table3}, we present the raw (without any compression) number of floating points we need to communicate through the channel for all three methods. Here, $C$ represents the number of floating points in the shared conditions which is common to all the methods. $E$ represents the size of the edge map (binary map), which counts the additional conditions introduced in this work. Compared to OD, Diff-GO can save up to 0.5 M floating points due to novel quantized noise space. The results imply the capacity of our method to reconstruct the same meaningful data while saving GO communication costs. Table~\ref{tab:table1} demonstrates the benefits of encoding random vectors via prior-selected bases.
\vspace{-0.1cm}
\begin{table}[h]
\caption{Semmantic similarity of the generated images with different methods evaluated against different metrics: LPIPS and FID. Some of the results are from the work GESCO~\cite{babarosa} (Smaller value is better). We present different choices of $n$ for our proposed Diff-GO. } 
\label{tab:table1}
\centering
\begin{tabular}{|c||c||c|c|}
\hline
Method & LPIPS$\downarrow$ & FID$\downarrow$ \\
\hline
SPADE~\cite{SPADE}& 0.546 & 103.24\\
\hline
CC-FPSE~\cite{CC-FPSE}& 0.546 & 245.9\\
\hline
SMIS~\cite{SMIS}& 0.546 & 87.58\\
\hline
OASIS~\cite{OASIS}& 0.561 & 104.03\\
\hline
SDM~\cite{SDM}& 0.549 & 98.99\\
\hline
OD& 0.2191 & 55.85\\
\hline
GESCO& 0.591 & 83.74\\
\hline
RN &  0.3448 & 96.409\\
\hline
Diff-GO (n=20)& 0.3206 & 74.09\\
\hline
Diff-GO (n=50)& 0.2697 & 72.95\\
\hline
Diff-GO (n=100)& 0.2450 & 68.59\\
\hline

\end{tabular}
\end{table}
\vspace{-0.5cm}
\subsection{Evaluation in Downstream Tasks}
Next, we measure the performance of the proposed framework in downstream tasks.
We first present the results of the object detection. For the evaluations, we use the pre-trained object detection model from~\cite{dtr}. In this experiment, we evaluate the mean intersection over union (mIoU) for three different objects of interest. From Table~\ref{tab:table2}, Diff-GO achieves superior performance similar to OD, while the GESCO underperforms. Note that, here, we use forward diffusion to derive the best noise for all the methods. 

We also evaluate the performance in depth map estimation as presented in
in Table~\ref{tab:table2}. Here evaluate the depth estimation using root mean square error (RMSE) between ground truth and reconstructed image. For depth map generation from street images, we use a pre-trained model from the work~\cite{depth}. From Table~\ref{tab:table2}, Diff-GO performs the best, in comparison with OD and GESCO, which show the promising power of Diff-GO for downstream tasks of depth sensing in autonomous driving.

\subsection{Ablation Study of Different Size of Noise Space}
In this part, we experiment with different numbers of noise basis vectors to assess their impact. We use 100 noise vectors and examine FID, LPIPS, SAMSR, and SAMSS scores for sharing various quantities, denoted as $n_i$. We choose $n_i = 1, 10, 20, 50,$ and $100$. From Table~\ref{tab:semscore}, FID and LPIPS scores decrease as we share more basis vectors, while SAMSR and SAMSS scores increase with $n$. We also demonstrate the flexibility of using downstream tasks (death estimation) for determining the number of weights to share with the receiver which follows the same trend. As shown in Table~\ref{tab:semscore}, even sharing just one essential weight can lead to efficient message regeneration and communication cost savings.
\begin{table}[h]
\caption{Object detection: mean intersection over the union (mIoU$\uparrow$) of objects of interest. We use pre-trained model form~\cite{dtr} for object detection on the generated images and error of depth estimation in RMSE. } 
\label{tab:table2}
\centering
\begin{tabular}{|c||c|c|c|c|}
\hline
Method & Car & People & Bicycle & Depth (RMSE$\downarrow$) \\
\hline
GESCO & 68.27 & 62.89 & 62.26 & 0.1489\\ 
\hline
OD & 73.52 & \textbf{68.96} & 67.30 & 0.1183\\ 
\hline

Diff-GO 100 & \textbf{73.55} & 67.57 & \textbf{67.74}& \textbf{0.1077} \\ 
\hline

\end{tabular}
\vspace{-3mm}
\end{table}


\begin{table}[h]
\vspace{-0.1cm}
\caption{GO-QOS of reconstructed images in terms of SAM score, and transmitted floating points to the receiver end by three different approaches. (SAMSR: SAM score with respect to the ground truth; SAMSS: SAM score with respect to the segmentation map of the ground truth.)} 
\label{tab:table3}
\centering
\begin{tabular}{|c||c|c|c|}
\hline
Method  & SAMSR$\uparrow$ & SAMSS$\uparrow$ & Transmitted \\
&  & & Floating Points$\downarrow$ \\
\hline
GESCO & 0.9738&0.9744&C\\ 
\hline
OD &\textbf{0.9941}&0.9839&C+E+0.5M\\ 
\hline
Diff-GO 100 &0.9927&\textbf{0.9864}&C+E+100\\ 
\hline

\end{tabular}
\vspace{-2mm}
\end{table}


\begin{table}[h]
\vspace{-0.2cm}
\caption{How different QoS scores vary with the number of basis vectors of the noise latent. Depth represents the depth estimation as GO-QOS score. Here, $n_1 =1, n_3 = 10, n_4 = 20, n_5 = 50, n = 100$} 
\label{tab:semscore}
\centering
\begin{tabular}{|c||c|c|c|c|c|c|}
\hline
Metric & 1  & 10 & 20 & 50 & 100\\
\hline
FID$\downarrow$& 70.252  & 69.43 & 69.45 & 69.46 & 68.59\\ 
\hline
LPIPS$\downarrow$& 0.2727 & 0.2794 & 0.2793 & 0.2793 & 0.2450\\ 
\hline
SAMSR$\uparrow$ &0.98611 & 0.9861 & 0.9861 & 0.9861 & 0.9861\\ 
\hline
SAMSS$\uparrow$&0.9854  & 0.9854 & 0.9854 & 0.9854 & 0.9927\\ 
\hline
Depth (RMSE)$\downarrow$&0.1078  & 0.1077 & 0.1077 & 0.1077 & 0.1077\\ 
\hline
\end{tabular}
\end{table}

\vspace{-0.4cm}
\section{Conclusions and Future Works}


This work introduces a novel goal-oriented (GO) communication system that generalizes the concept of semantic communications and prioritizes the downstream tasks that rely on
the communicated signals to trade computation for bandwidth
efficiency (C4BE). The Diff-GO communication design achieves
ultra-high bandwidth efficiency by utilizing generative AI 
at its core. Unlike many semantic
communication works using generative AI, our approach
considers the goal of communications beyond the basic language model (or semantics).  We leverage a forward diffusion process and 
use a unique low-dimensional noise space for bandwidth reduction. To ensure goal-oriented QoS (GO-Qos), our approach implements local generative feedback without altering any existing communication links and protocols.  Future research will explore different noise spaces for improved performance and consider developing metrics for measuring GO-QoS.

\bibliographystyle{IEEEtran}
\bibliography{IEEEabrv,user_scheduling}{}

\end{document}